\documentclass[sigconf, nonacm=true]{acmart}

\usepackage[textsize=tiny,textwidth=1.5cm]{todonotes}
\usepackage[font=small,labelfont=bf]{caption, subcaption}
\usepackage{siunitx}
\usepackage{enumerate}
\usepackage{multirow}
\usepackage{pgfplots}
\usepackage{fancyhdr}
\usepackage{hyperref}
\hypersetup{
    colorlinks,
    linkcolor={red!50!black},
    citecolor={green!45!black},
    urlcolor={blue!80!black}
}

\AtBeginDocument{%
  \providecommand\BibTeX{{%
    \normalfont B\kern-0.5em{\scshape i\kern-0.25em b}\kern-0.8em\TeX}}}

\begin{document}

\title{Are you still with me? Continuous Engagement Assessment from a Robot's Point of View}
\author{Francesco Del Duchetto}
\affiliation{%
   \institution{\emph{L-CAS, University of Lincoln}}
   \city{Lincoln}
   \state{UK}
}
\email{fdelduchetto@lincoln.ac.uk}

\author{Paul Baxter}
\affiliation{%
   \institution{\emph{L-CAS, University of Lincoln}}
   \city{Lincoln}
   \country{UK}
}
\email{pbaxter@lincoln.ac.uk}

\author{Marc Hanheide}
\affiliation{%
   \institution{\emph{L-CAS, University of Lincoln}}
   \city{Lincoln}
   \country{UK}
}
\email{mhanheide@lincoln.ac.uk}

\begin{abstract}
  Continuously measuring the engagement of users with a robot in a Human-Robot Interaction (HRI) setting paves the way towards \emph{in-situ} reinforcement learning, improve metrics of interaction quality, and can guide interaction design and behaviour optimisation.
  However, engagement is often considered very multi-faceted and difficult to capture in a workable and generic computational model that can serve as an overall measure of engagement.
  Building upon the intuitive ways humans successfully can assess situation for a degree of engagement when they see it, we propose a novel regression model (utilising CNN and LSTM networks) enabling robots to compute a single scalar engagement during interactions with humans from standard video streams, obtained from the point of view of an interacting robot. 
  The model is based on a long-term dataset from an autonomous tour guide robot deployed in a public museum, with continuous annotation of a numeric engagement assessment by three independent coders. 
  We show that this model not only can predict engagement very well in our own application domain but show its successful transfer to an entirely different dataset (with different tasks, environment, camera, robot and people). The trained model and the software is available to the HRI community as a tool to measure engagement in a variety of settings.
\end{abstract}




\maketitle

\thispagestyle{fancy}
\fancyhf{}
\lhead{\textbf{Published in \emph{Frontiers in Robotics and AI}, vol. 7, Sep. 2020, \href{https://doi.org/10.3389/frobt.2020.00116}{https://doi.org/10.3389/frobt.2020.00116}. }}

\renewcommand{\headrulewidth}{0.7pt}

\section{Introduction}\label{sec:intro}

\begin{figure}[t]
    \centering

    \includegraphics[width=\linewidth]{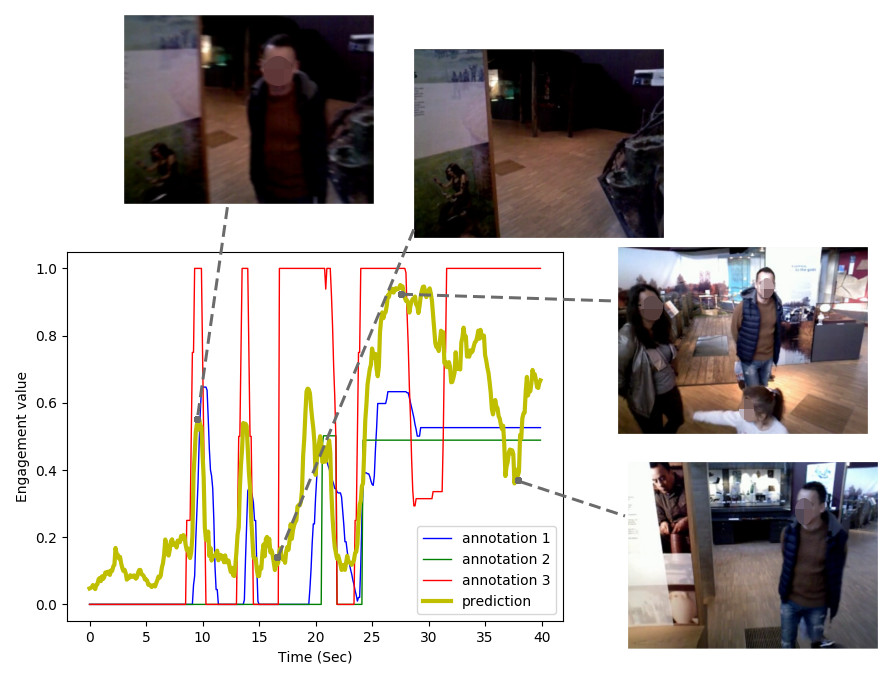}

    \caption{Engagement annotated values and our model's predictions over a guided tour interaction sequence recorded from our robot's head camera.}
    \label{fig:comparison}
\end{figure}

One of the key challenges for long-term interaction in human-robot interaction (HRI) is to maintain user engagement, and, in particular, to make a robot aware of the level of engagement humans display as part of an interactive act. 
With engagement being an inherently internal mental state of the human(s) interacting with the robot, robots (and observing humans for that matter) have to resort to the analysis of external cues (vision, speech, audio).

%
In the research program that informed the aims of this paper, we are working to close the loop between the user perception of the robot as well as their engagement with it, and our robot's behavior during real-world interactions, i.e., to improve the robot's planning and action over time using the responses of the interacting humans.
The estimation of users' \emph{engagement} is hence considered an important step in the direction of automatic assessment of the robot's own behaviours in terms of its social and communicative abilities, in order to facilitate \emph{in-situ} adaptation and learning. 
In the context of reinforcement learning, a scalar measure of engagement can directly be interpreted as a \emph{reinforcement signal} that can eventually be used to govern the learning of suitable actions in the robot's operational situation and environment. 
As a guiding principle (and indeed a working hypothesis), we anticipate that higher and sustained engagement with a robot can be interpreted as a positive reinforcement of the robot's action, allowing it to improve its behavior in the long term.

Previous work on robot deployment in museum contexts~\cite{Duchetto2019} provide evidence on how user engagement during robot guided tours easily degrades with time when employing an open-loop interactive behavior which does not take into account the engagement state of the other (human) parties. 

However, we argue that the usefulness of a scalar measure of engagement as presented in the paper stretches far beyond our primary aim to use it to guide learning. 
Work in many application domains of HRI~\cite{Rudovic2017, Baxter2014, Ben-Youssef2017} has focused on a measure of engagement to inform the assessment of the implementation for a specific use-case, or to guide a robot's behavior. 
However, how engagement is measured and represented varies greatly (see Sec.~\ref{sec:related}) and there is yet to be found a generally applicable measure of engagement that readily lends itself to guide the online selection of appropriate behavior, learning, adaptation, and analysis.
Based on the observation that engagement as a concept is implicitly often quite intuitive for humans to assess, but inherently difficult to formalize into a simple and universal computational model, we propose to employ a data-driven machine learning approach, to exploit the implicit awareness of humans in assessing an interaction situation.
Consequently, instead of aiming to comprehensively model and describe engagement as a multi-factored analysis, we use end-to-end machine learning to directly learn a regression model from video frames onto a scalar in the range of $0\%$ to $100\%$, and use a rich annotated dataset obtained from a long-term deployment of a robot tour guide in a museum to train said model.

For a scalar engagement measure to be useful in actual HRI scenarios, we postulate that a few requirements have to be fulfilled. In particular, the proposed solution should
\begin{itemize}
    \item demonstrably generalize to new unseen people, environments, and situations;
    \item operate from a robot's point of view, forgoing any additional sensors in the environment;
    \item employ a sensing modality that is readily available on a variety of robot platforms; 
    \item have few additional software dependencies to maximize community uptake; and
    \item operate with modest computational resources at soft real-time.
\end{itemize}

Consequently, we present our novel engagement model, solely operating on first-person (robot-centric) point of view video of a robot and prove its applicability not only in our own scenario but also on a publicly available dataset (UE-HRI) without any transfer learning or adaptation necessary.
We demonstrate that the model can operate at typical video frame rates on average GPU hardware typically found on robots.
Hence, the core contributions of this paper can be summarised as
\begin{enumerate}[i]
    \item the appraisal of a scalar engagement score for the purpose of in-situ learning, adaptation, and behavior generation in HRI;
    \item a proposed end-to-end deep learning architecture for the regression of first-person view video stream onto scalar engagement factors in real-time; 
    \item the comprehensive assessment of the proposed model on our own long-term dataset, and a publicly available HRI dataset proving the generalizing capabilities of the learned model; and
    \item the availability of a implementation and trained model to provide the community with an easy to use, out of the box methodology to quantify engagement from first-person view video of an interactive robot.
\end{enumerate}

\section{Assessment of Engagement}~\label{sec:related}



Recognizing the level of engagement of the humans during the interactions is an important capability for social robots. In the first place, we want to recognize the level of engagement as a way to assess the robot behavior. Feeding this information to a learning system we can improve the robot behavior to maximize the level of engagement. In an education scenario, such as 
a museum, being able to engage the users is a crucial factor. It is known that higher level of engagement generates better  learning outcomes \cite{ponitz2009kindergarten}, while engagement with a robot during a learning activity has also been shown to have a similar effect \cite{gleason2017hybrid}. While there is evidence that the presence of a robot, particularly when novel, is sufficient in itself for higher engagement in educational STEM activities, e.g. \cite{baxter2018engaging}, the focus in the present work is on engagement between individuals and the robot within a direct (social) interaction, for which there is not a universally agreed definition \cite{Glas2015}.

Within interactions, engagement has been characterized as a process that can be separated in four stages: point of engagement, period of sustained engagement,  disengagement, and re-engagement \cite{o2008user}. Context has also been identified as being of importance, in terms of the task and environment, as well as the social context \cite{Castellano2012}. For example, \cite{michalowski2006spatial} proposes a simple model to infer engagement for a robot receptionist based on the person spatial position within some predefined areas around the robot, and \cite{salam2015engagement} studies to what extent is possible to predict the engagement of an entity relying solely on the features of the other parties of the interaction, showing that engagement, and the features needed to detect it, changes with the context of the interaction \cite{salam2015multi}.
These examples furthermore suggest that there are multiple, overlapping, and likely interacting timescales involved in the characterization of engagement, from the longer term context to short interaction-orientated behaviours that nevertheless impact social dynamics, and which humans are particularly receptive to \cite{durantin2017social}.

In the context of the characterization of engagement above, there are a number of approaches to the automatic assessment of engagement that may be distinguished. On the one hand, there is a focus on individual behavioral cues, which may be integrated to form a characterization of engagement. On the other hand, there is a more holistic perspective of engagement taken, where proxy metrics may be used or direct measures of engagement estimated. Combinations of these perspectives are summarised briefly below.

Work on characterizing engagement in both human-human and human-robot interactions has identified human gaze as being of particular significance when determining engagement levels in an interaction, e.g. \cite{rich2010recognizing,holroyd2011generating}. Gaze thus forms an important behavioral cue when assessing engagement, e.g. \cite{sidner2004look,Baxter2014}. For example, Lemaignan et al. \cite{lemaignan2016real} do not try to directly define and detect engagement, recognizing that it is a complex and broad concept. Instead, the concept of ``with-me-ness'' is introduced, which is the extent to which the human is ``with'' the robot during the interactions, and which is based on the human gaze behavior. 

Beyond only human gaze behavior, Foster et al. \cite{foster2017automatically}, for example, address the task of estimating the engagement state of customers for a robot bartender based on the data from audiovisual sensors. They test different approaches reporting that the rule-based classifier shows competitive performances with the trained ones and could actually be preferred for their stability (and to overcome data-scarcity problems).


In addition to these explicitly cue-centred approaches, more recently, attempts have been made to leverage the power of machine learning to discover the important overtly visible features with minimal (or at least sparse) explicit guidance from humans (through cue identification for example). 
For example, \cite{won2019personalized} use an active learning approach with Deep RL to automatically (and interactively) learn the engagement level of children interacting with a robot from raw video sequences. The learning is incremental and allows for real-time update of the estimates, so that the results can be adapted to different users or situations. The DQN is initially trained with videos labeled with engagement values. In other work \cite{rudovic2018culturenet} investigate  the  performance of  deep  learning  models  in  the  task  of  automated  engagement estimation  from  face  images  of  children  with  autism using  a  novel  deep  learning  model,  named  CultureNet, which efficiently leverages the multi-cultural data when performing  the  adaptation  of  the  proposed  deep  architecture to the target culture and child, although this is based on a dataset of static images rather than real-time data. 

These deep learning methods have the advantage that the constituent features of interest do not have to be explicitly defined \textit{a priori} by the system designer, rather, only the (hidden) phenomenon needs to be annotated; engagement in this case. Since social engagement within interactions is readily recognized by humans based on visible information (see discussion above), human coding of engagement provides a promising source of ground-truth information. Indeed, in this context, \cite{tanaka_socialization_2007} employed human coders to assess the `quality' of observed interactions, demonstrating good agreement between coder on what was a subjective metric.

Taken together, the literature indicates that while a precise operational definition of engagement may not be universally agreed, it seems that more holistic perspectives may be more insightful. It is likely that while gaze is an important cue involved in making this assessment, there are other contextual factors that influence the interpretation of engagement. Given that humans are naturally able to accurately assess engagement in interactions, it seems that one promising possibility would be to leverage this to directly inform automated systems.

\section{Preliminaries}
This work is embedded in a research program that seeks to employ online learning and adaptation of an autonomous mobile robot to deliver tours in a museum context. The robotic platform, described below, has been operating autonomously in this environment for an extended period of time, as evidenced by the long term autonomy metrics (Table \ref{tab:metrics}).

The goal is to facilitate the visitor's engagement with the museum's display of art and archaeology.
This project provides an opportunity to study methodologies to equip the robot with the ability to interact socially with the visitors. In particular, the research aims to find a good model to allow the robot to do the correct thing at the right moment, in terms of social interaction. The first step in doing so is endowing the robot with a means of assessing its own performance at any given moment to allow adaptation, learning, and to avoid repeating the same errors.

\subsection{Robotic Platform}
The robot is a Scitos G5 robot manufactured by MetraLabs GmbH. It is equipped with a laser scanner with 270$^{\circ}$ scan angle on its base and two depth cameras. An Asus xtion depth camera is mounted on a pan-tilt unit above his head and a Realsense D415 is mounted above the touchscreen with an angle of 50$^{\circ}$ w.r.t. the horizontal plane in order to face the people standing in front of the robot. The interactions with the visitors are mediated through a touch screen, two speakers, a microphones array and a head with two eyes that can move with five degrees of freedom to provide human-like expressions. To ensure safe operations in public environments the robot is equipped with an array of bumpers around the circular base with sensors to detect collisions and two easily reachable emergency buttons that, when activated, cuts the power to the motors.
The software framework is based on ROS and uses STRANDS project \cite{hawes2017strands} core modules for topological navigation, people tracking, task scheduling and data collection.

\subsection{Long Term Deployment Analysis}
The data gathered so far spans the date range between the 24$^{\text{th}}$ January 2019 (day on which we started recording data of the robot operations) and the 9$^{\text{th}}$ May 2019, with data collection remaining ongoing.
The work and data recording exercise has been approved by the University of Lincoln's Ethics Board, under approval ID "COSREC509". The ethical approval does not allow the public release of any data that can feature identifiable persons, in particular video data.



\begin{table}
\vspace{0.2cm}
\caption{Long-Term Autonomy metrics: \emph{total system lifetime} (TSL - how long the system is available for autonomous operation), and \emph{autonomy percentage} (A\% - duration the system was actively performing tasks as a proportion of the time it was allowed to operate autonomously), following \cite{hawes2017strands}.}
\label{tab:metrics}
\centering
\begin{tabular}{|l|c|}
\hline
{Days of operation} & 103 days \\ \hline
{Total distance travelled} & 299 km \\ \hline
{Total tasks completed} & 8423 \\ \hline
{TSL} & 26 days, 11 hours \\ \hline
{A\%} & 74\% \\ \hline
\end{tabular}
\end{table}


During the current deployment the robot performs mainly two types of interactive task: \textit{guided tour} and \textit{go to exhibit and describe}. In the first task the robot guides the users to 5 or 6 exhibits sequentially around the museum, describing what they contain when stopping in front of each. During the second interactive task the robot guides to users to one of the exhibits and, when arrived at the destination, describes the content it is showing.

\subsection{TOGURO Dataset Collection}\label{sec:toguro}

The TOur GUide RObot (TOGURO) dataset was collected from the two cameras mounted on the robot's body and head, each providing a stream of rgb and depth frames. These video streams were collected from the start until the end of each \textit{guided tour} and \textit{go to exhibit and describe} task. Considering the large number of videos to be stored each we saved the frame streams as compressed MPEG video files directly from the interaction. Moreover, we store, in an additional file, the ROS timestamp at the time each frame is received by the video recorder node. This allows us to reconstruct afterward the alignment between the different video streams frame by frame.

The participants were aware that the robot was recording data during the interactions (by means of visible signs and leaflets), although they were not informed that the purpose of this data was for engagement analysis, thus not biasing their behaviours.
In total we collected 703 distinct interactions 
with a total duration of 40 hours and 17 minutes. As described below (section \ref{sec:data_coding}), only a subset of this total data was coded.
Given the unconstrained setting, the interactions varied significantly in duration, with the shortest at 1.2 seconds and the longest at 2 hours 40 minutes.

Given that the museum in which the robot is deployed is a public space openly accessible to anyone, the interactions between the robot and the museum's visitors are completely unstructured. People walking in the gallery are allowed to roam around the collection or to interact with the robot. When they choose to do so they do not receive any instruction about how to interact with it explicitly, and are not observed by experimenters when doing so.

\subsection{Dataset Coding}~\label{sec:data_coding}
In order to address the primary research goal -- the assessment of robot-centric group engagement -- the dataset was manually coded in order to establish a ground truth. As noted previously, given that there is not a universally accepted operationalized definition of engagement, a human observer response method is employed in the present work, following the prior application of a continuous audience response method \cite{tanaka_socialization_2007}.

\begin{figure}
    \centering
    \includegraphics[width=0.8\linewidth]{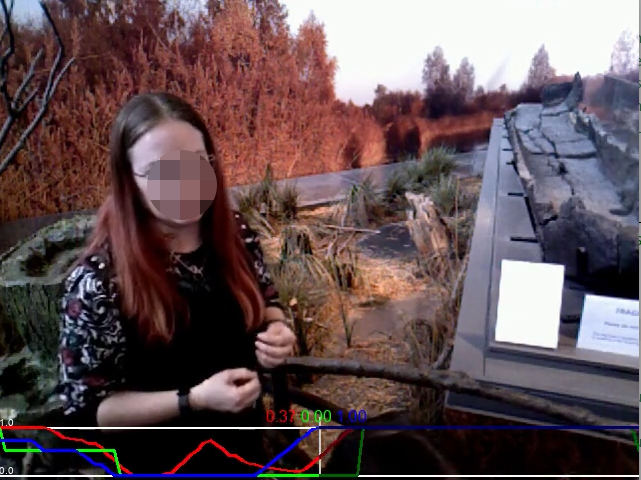}
    \caption{One frame from a video in the TOGURO dataset recorded from the robot's head camera during a guided tour. The red, green and blue plots at the bottom of the frame represent each a distinct annotation sequence.}
    \label{fig:video_ann}
\end{figure}

The annotations were performed over only the rgb stream of robot's head camera, and not taking into account all the four video streams available from the collected data.
Similarly to \cite{tanaka_socialization_2007}, the annotators were asked to indicate in real time how engaged people interacting with a robot appeared to be in a video captured by the robot (e.g. Figure \ref{fig:video_ann}). They operated a dial using a game-pad joystick while watching the interaction videos using the NOVA annotation tool\footnote{https://github.com/hcmlab/nova} \cite{baurcontext2015}. This procedure allowed the generation of per-frame annotations of the provided videos, with very little time spent on software training (around 20 minutes per annotator) and on the annotation process itself (not more than the duration of the videos). The annotators were instructed by providing them with a demonstration, and a set of annotation rules based on a set of typical examples\footnote{Available at: https://justpaste.it/6p1tb/pdf}.

Three annotators took part in the coding process: each was familiar with the robot being used and the interaction context. Three subsets of the overall dataset collected were randomly drawn and assigned to the annotators. The subsets were partially overlapping. This was to enable an analysis of inter-rater agreement to assess reliability of the essentially subjective metric, but also to maximize annotation coverage of the dataset. As indicated in Table \ref{tab:ann_data}, the total length of the annotated data was over nine hours, with 3 hours 27m of overlap between the annotators (resulting in 5 hours 50m of unique videos annotated).

The amount of annotated data is depicted in Table \ref{tab:ann_data}. 96 unique videos were coded by the three annotators with a total of 146 videos (including repeated annotations) for a total duration of 9 hours and 17 minutes. In total the annotated video set features 227 people (53.74\% (122) females and 46.26\% (105) males, 60.79\% (138) adults and 39.21\% (89) minors). The composition of each group of people interacting with the robot is very diverse; on average each videos features $2.41$ people ($min=0, max=9, \sigma=1.56$), $1.32$ females ($min=0, max=6, \sigma=0.89$), $1.14$ males ($min=0, max=5, \sigma=1.26$), $1.5$ adults ($min=0, max=5, \sigma=0.97$) and $0.96$ minors ($min=0, max=6, \sigma=1.14$).

\begin{table}
  \caption{Video annotations by annotator (coder): unique indicates length of video coded by a single coder}
  \label{tab:ann_data}
  \begin{tabular}{ccc}
    \toprule
    Coder & \# Videos & Tot Duration\\
    \midrule
    \texttt{Coder1} &  \num{66}  & 3h 59m\\
    \texttt{Coder2} &  \num{40}  & 2h 55m\\
    \texttt{Coder3} &  \num{40}  & 2h 23m\\
  \bottomrule
    Unique & \num{94} & 5h 50m\\
    Total & \num{146}  & 9h 17m\\

  \bottomrule
\end{tabular}
\end{table}

\begin{figure}
    \centering
    \includegraphics[width=\linewidth]{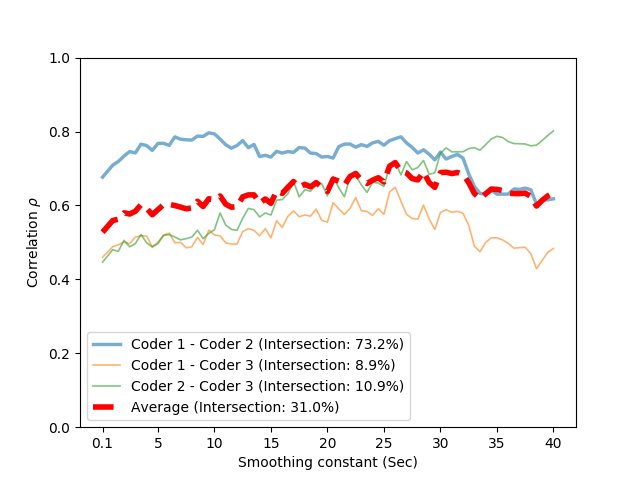}
    \caption{Spearman correlation averaged over coder pairs and weighted by the overlap rate. Value reported over different smoothing constants $S$.}
    \label{fig:correlation}
\end{figure}

The annotated engagement rating is a continuous scalar for every frame of video data. As such, Spearman's rank correlation ($\rho$) is employed to assess inter-rater agreement. 
Table \ref{tab:corr_values} shows the correlation values for each pair of annotators. Since every frame is annotated (with a frame-rate of \num{10} frames-per-second), the continuous values were smoothed over time, using different smoothing constant values, in the range $[0.1s, 40s]$ (Figure \ref{fig:correlation}). Table \ref{tab:corr_values} provides a summary of these, with overall mean agreement rates at selected representative values of the smoothing constant. While there is some variability in the between-coder agreement, mean values of $\rho$ vary in strength from moderate to strong (0.56 to 0.72). In this regard, there is a trade-off to be made between the smoothing constant size and the apparent agreement between the coders: the larger time window size reduces the real-time relevance of the engagement assessment, even though the agreement over the extended periods of time is greater than in comparatively shorter windows. Overall, these results indicate that the use of the independently coded data can be considered reliable in terms of the highly variable and subjective metric of engagement.

\begin{table}
  \caption{Spearman's Correlation $\rho$ at different smoothing constant values $S$. The significance $p$-value $< 0.001$ and sample size $n \geq 89$ for all coder pairs and smoothing constants.}
  \label{tab:corr_values}
  \begin{tabular}{c c c}
    \toprule
    Coders Pair & $S$ (sec) & $\rho$ \\
    \midrule
    \multirow{4}{*}{
        \texttt{Coder1} $\leftrightarrow$ \texttt{Coder2}} & 1 & 0.71  \\
        & 5 & 0.77\\
        & 10 & 0.79 \\
        & 26 & 0.78\\
    \hline
    \multirow{4}{*}{
        \texttt{Coder1} $\leftrightarrow$ \texttt{Coder3}} & 1 & 0.49 \\
        & 5 & 0.5\\
        & 10 & 0.52\\
        & 26 & 0.65\\
    \hline
    \multirow{4}{*}{
        \texttt{Coder2} $\leftrightarrow$ \texttt{Coder3}} & 1 & 0.48 \\
        & 5 & 0.5\\
        & 10 & 0.53\\
        & 26 & 0.72\\
    \bottomrule
    \multirow{4}{*}{Average} & 1 & 0.56\\
        & 5 & 0.59\\
        & 10 & 0.62\\
        & 26 & 0.72\\
  \bottomrule
\end{tabular}
\end{table}

\section{The engagement regression model}~\label{sec:eng_model}

\begin{figure*}
    \centering
    \includegraphics[width=0.8\textwidth]{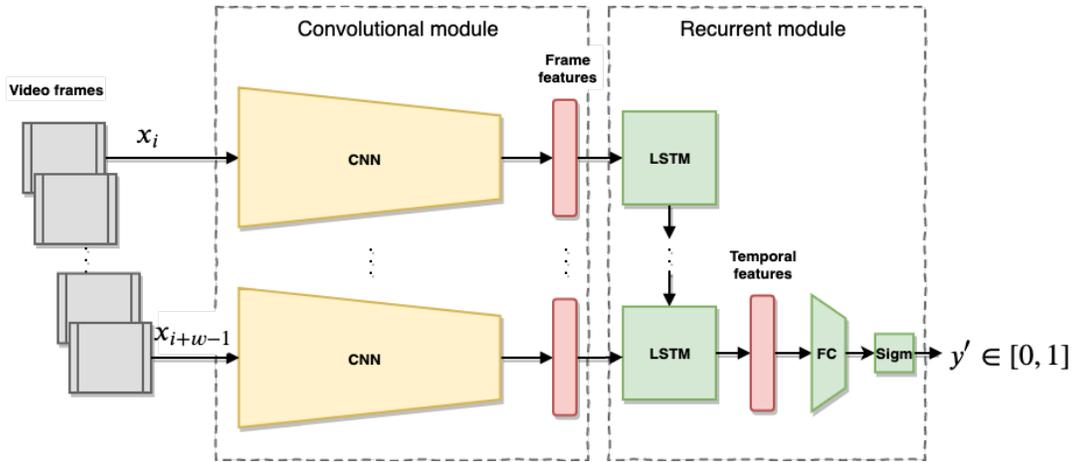}
    \caption{Overview of the proposed model. The input is a video stream of interactions between the robot and humans collected in $w$ size intervals. The frames $x_i$ are passed through the pre-trained CNN (ResNet) producing a per-frame feature vector which is then passed sequentially to the LSTM network. After $w$ steps the LSTM produces a temporal feature vector which is passed to a FC layer with sigmoid activation to produce an engagement value $y$ for the temporal window. }
    \label{fig:model}
\end{figure*}

Given the ground-truth provided by the human-coded engagement levels within interactions with the robot, we propose a deep learning approach for the estimation of human engagement from video sequences. The model is trained end-to-end from the raw images coming from the robot's head camera to predict a high-level engagement score of people interacting with the robot. 
It should be noted that this model does not model individual humans in the view of the robot but provides an overall holistic engagement score.

The network architecture, depicted in Figure \ref{fig:model}, is composed of two main modules: a convolutional module which extracts frame-wise image features and a recurrent module that aggregates the frame features over a time to produce a temporal feature vector of the scene. 
The convolutional module is a ResNetXt-50 Convolutional Neural Network (CNN) \cite{xie2017aggregated} pre-trained on the ImageNet dataset \cite{krizhevsky2012imagenet}. We obtain the frame features from the activation of the last fully connected layer of the CNN, with dimension 2048, before the softmax layer. 
The recurrent module is a single layer Long-Short Term Memory (LSTM) \cite{hochreiter1997long} with $2048$ units followed by a Fully Connected (FC) layer of size $2048 \times 1$. The LSTM takes in input a sequence of $w$ frame features coming from the convolutional module and produces in turn a feature vector that represents the entire frame sequence, to capture temporal behavior of humans within the time window $w$.
The temporal features are passed through the FC layer with a sigmoid activation function at the end to produce values $y' \in [0,1]$
The recurrent module is trained in our experiments to predict engagement values from the provided annotation values, while the CNN layer is fixed.


The proposed framework is implemented in Python using the Keras library~\cite{chollet2015keras} and will be freely released as a ready-to-use tool to the HRI community.



\section{Experiments}
We train and test the model presented in Section \ref{sec:eng_model} on our own TOGURO dataset, and assess generalization of this model (without modification) on the public UE-HRI dataset in the following subsections.

\subsection{TOGURO Dataset Processing}\label{sec:TOGURO}


We used the entire annotated dataset presented in Section \ref{sec:data_coding}, composed of 94 videos, for a total duration of 5 hours and 50 minutes of interactions.
For each video we randomly choose an annotation, if multiple are available from the different coders (see table~\ref{tab:ann_data}), in order to avoid repetitions in the data and biasing the model toward those videos that have been annotated multiple times.
Each video is then randomly assigned to either the training, test or validation set with a corresponding probability of 50\%, 30\% and 20\%, respectively, to prevent our model to train and test over data that are closely correlated at the video frames level. Sampling for the dataset split hence operates on full video level, rather than on frame level.
Each video $V_k$ is composed of $I_{V_k}$ frames $x_i \in V_k$ for $i \in 0, \dots , I_{V_k}$ and has an associated array of annotations $A_k = [y_0, \dots , y_{I_{V_k}}]$, also of dimension $I_{V_k}$.
From all the videos in each set (training/test/validation) we extract all the possible sequences of $w$ consecutive frames $X_i = [x_i, \dots , x_{i+w-1}]$ to be the input sample for our model. 
Therefore, each sample $X_i$ has an overlap of $w-1$ frames with the consecutive sample $X_{i+1}$ from the same video.
For each sample $X_i$ we assign the ground truth value $y_{i+w-1} \in A_k$, in order to relate each sequence of frames with the engagement value set at the end of the sequence.

After the pre-processing phase over our dataset we obtain \num{93271} training samples, \num{72146} test samples and \num{44581} validation samples.
Each frame is reshaped to $224 \times 224$ pixel frames, and normalized before being fed to the network. 

\subsection{Training and Evaluation}
For training and evaluation we decided to set the window size $w$ equal to 10 frames in order to have a model that gives evaluations of the engagement in a relative short times (i.e. after 1 second). Even
though more temporally extended time windows would provide more coherent ground truth values among the different annotators, as discussed in Section \ref{sec:data_coding}, we decide to sacrifice some accuracy in favour of increase realtimeness of our model predictions.

During training the weights of the Convolutional module, which is already pre-trained, are kept frozen while the Recurrent module is fully trained from scratch.
The model is trained to optimize the Mean Squared Error (MSE) regression loss between the prediction values $y'_i$ and the corresponding ground truth values $y_i$ using the Adagrad optimization algorithm \cite{duchi2011adaptive} with an initial learning rate $lr = 1e-4$. At each training epoch we sample uniformly $20\%$ of the training set samples to be used for training and we collect them in batches of size $bs=16$. The uniform data sampling of the training data is performed in order to reduce training time and limiting overfitting \cite{el2019unrestricted}. 
The model has been trained for a total of 22 epochs using early stopping after no improvement on validation loss.



\subsection{Assessing generalization}\label{sec:ass_gen}

In order to assess the generalization capabilities of our trained model over different scenarios featuring people interacting with robots, we propose to test the performance of our trained model as a detector of the start and end of interactions over the UE-HRI dataset~\cite{Ben-Youssef2017}.
Similarly to our dataset, it provides video recordings from the robots own cameras allowing for engagement estimation from the robot's point of view.
The dataset provides videos of spontaneous interactions between humans and a Pepper (Softbank Robotics) robot alongside annotations of start/end of interactions and various signs of engagement decrease (Sign of Engagement Decrease (SED), Early sign of future engagement BreakDown (EBD), engagement BreakDown (BD) and Temporary Disengagement (TD)). 
The UE-HRI dataset features 54 interactions with 36 males and 18 females, where 32 are mono-users and 22 are multiparty.

For a fair comparison with our proposed method, 
we evaluate the ability of our model to distinguish between the moments during which an interaction is taking place and those in which there is a breakdown (TD or BD), the interaction is not yet started or it is already ended, in line with the UE-HRI coding scheme. 
Consequently, we predict engagement values over the RGB image streams from the Pepper robot's front camera.
By setting a threshold value $thr$ we convert the predictions $y'$ into a binary classification of $C = \{\top,  \bot\}$ (prediction above or below $thr$) which indicates whether there is engagement or not. The categorical predictions are then compared with values from the annotations in the dataset. We consider the ground truth value to be $y_{int}^t = \top$ if at time $t$ there is a annotation of a \emph{Mono} or \emph{Multi} interaction and there are no annotations of BD or TB in the UE-HRI coding.
The ground truth value is $y_{int}^t = \bot$ otherwise.

\section{Results}

\begin{table}
  \caption{Model performance on our TOGURO Dataset}
  \label{tab:model_performances}
  \begin{tabular}{cccc}
    \toprule
    GPU & Test loss & Prediction time & Mem. usage\\
    \midrule
    {\smaller GeForce GTX 1060} & \num{0.126} {\smaller(MSE)} & $t <= 0.05 sec$ & \num{5.4}GB \\
  \bottomrule
\end{tabular}
\end{table}
With our evaluation we set out to provide evidence that our model is able to predict engagement through regression on our own TOGURO dataset by assessing its accuracy in comparison to the ground-truth annotation, and to assess the generalization ability of the model on newly encountered situations through the analysis of the UE-HRI data. 

To show the ability of our framework to map short-term human behavioral features from image sequences into engagement scores, we compute the Mean Squared Error (MSE) prediction loss on our test set as $0.126$ (in the context of the $[0,1]$ interval of output expected), also reported in Table \ref{tab:model_performances}. 
Looking back at section~\ref{sec:intro}, soft real-time operation is seen as a requirement for the applicability of our model. Hence, we measured the duration of a forward pass on our GPU hardware of 10 consecutive frames (1 sample) through the the convolutional module and the recurrent module taking at most $50 ms$ (worst case), allowing real-time estimation of engagement at $20$ frames per second.    


\begin{figure}
    \centering
    \includegraphics[width=1\linewidth]{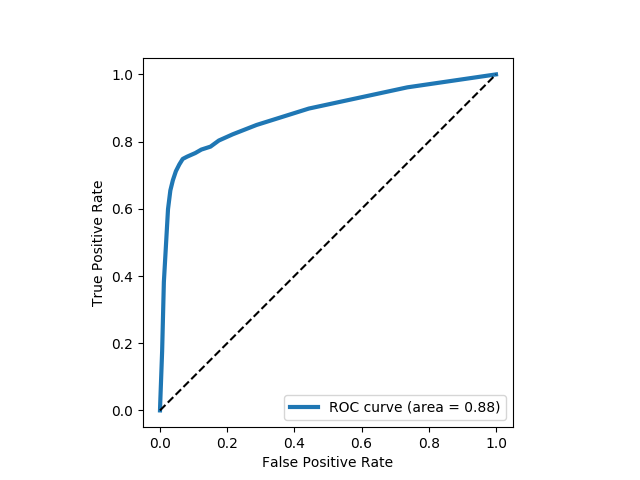}
    \caption{ROC curve generated using our trained model as a classifier of the interaction sessions for the UE-HRI dataset.}
    \label{fig:roc_curve10}
\end{figure}

Evaluating the power of our approach for binary classification on the UE-HRI as detailed above in section~\ref{sec:ass_gen}, allows us to capture the generalization capabilities.
In Figure \ref{fig:roc_curve10} we report the Receiver Operating Characteristic (ROC) curve obtained by varying the threshold with values in the range $thr \in [0, 1]$ of the binary classification task on the UE-HRI data. The Area Under the Curve ($AUC = 0.88$ in our experiment) reports the probability that our classifier ranks a randomly chosen positive instance $y_{int}^t = \top$ higher than a randomly chosen negative one $y_{int}^t = \bot$, i.e., provides a good assessment of the performance of the model in this completely different dataset.

\begin{figure*}
    \centering
    \includegraphics[width=\textwidth]{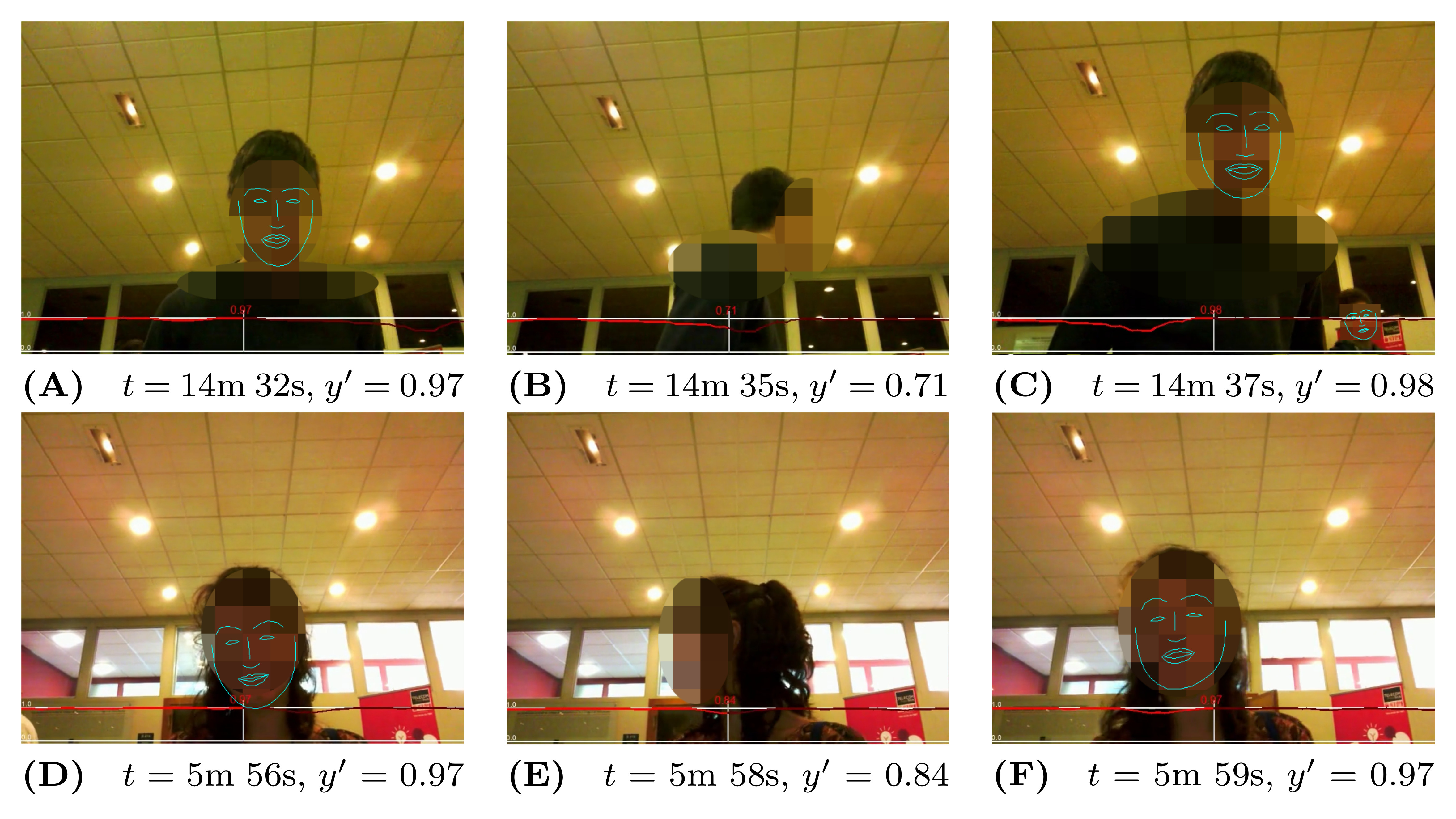}
    \caption{UE-HRI dataset: two sequences of short timescale sequential frames showing how the temporal diverting of attention is reflected in the model predicting a lower engagement value. Red plot shows the predicted engagement values over the frame sequences, with the prediction $y'$ at the frame shown in picture at time $t$ being in the center, past predictions on the left and future predictions on the right.}
    \label{fig:engagement_}
\end{figure*}

\begin{figure*}
    \includegraphics[width=\textwidth]{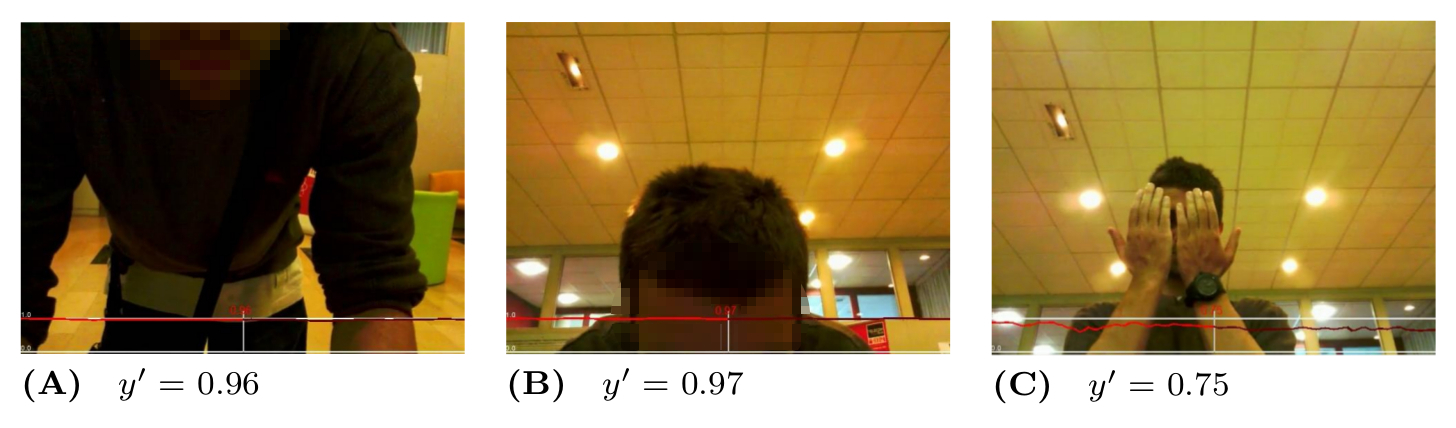}
    \caption{UE-HRI dataset: examples of correct prediction of high engagement ($y' >= 0.75$) in situations difficult to understand using standard face description features. Red plot shows the predicted engagement values over the frame sequences with the prediction $y'$ at the frame shown in picture being in the center, past predictions on the left and future predictions on the right.}
    \label{fig:difficult_predictions}
\end{figure*}

\begin{figure*}
    \includegraphics[width=\textwidth]{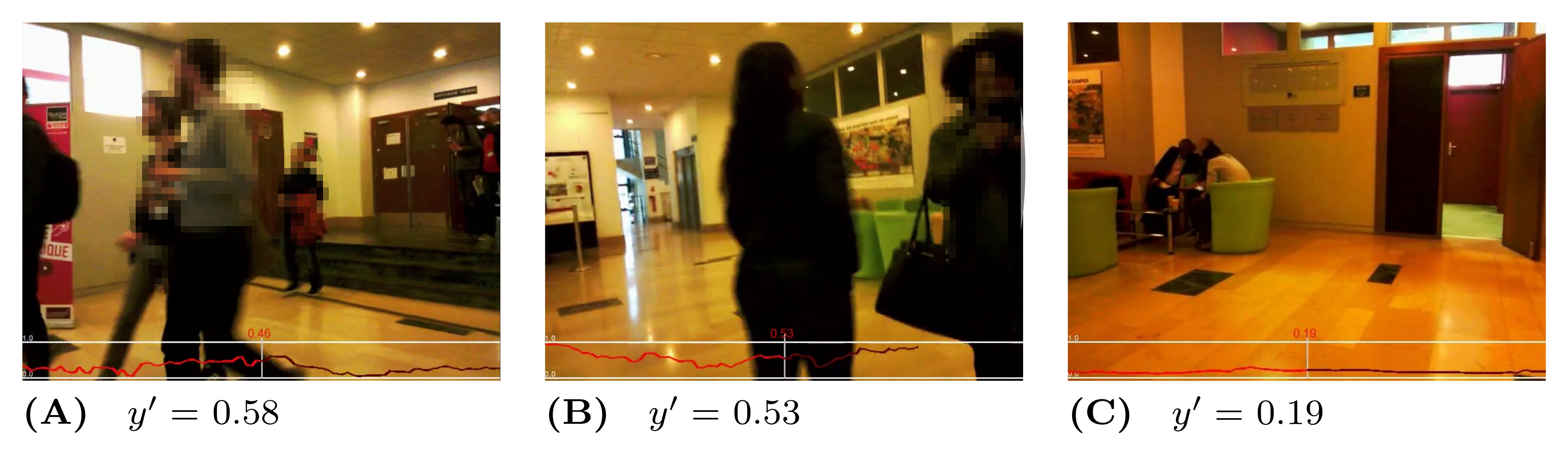}
    \caption{UE-HRI dataset: examples of correct low/medium engagement prediction ($y' <= 0.6$) in cases in which the people were not actually engaging with the robot. Red plot shows the predicted engagement values over the frame sequences with the prediction $y'$ at the frame shown in picture being in the center, past predictions on the left and future predictions on the right.}
    \label{fig:true_negative}
\end{figure*}

\section{Discussion and Conclusion}
This paper has motivated, developed and validated a novel easy-to-use computational model to assess engagement from a robot's perspective. The results presented in the previous sections lead us to the conclusion that
\begin{enumerate}[i]
    \item a moderate to strong inter-rater agreement (see table~\ref{tab:corr_values}) in measuring engagement on $[0,1]$ interval indicates that human can reasonably and reliably assess the holistic engagement from a robot's point of view solely from video;
    \item a two-stage deep-learning architecture as presented in figure~\ref{fig:model} trained from our TOGURO dataset is a suitable computational regression model to capture the inherent human interpretation of engagement provided by the annotators; and that
    \item the trained model is generic enough to be successfully applied in a completely different scenario, here the UE-HRI dataset, showing applicability of the model also in different environments, on a different robot with a different camera, and with different tasks and people. The area under the Receiver-Operator Curve (ROC) of $0.88$ in figure~\ref{fig:roc_curve10} evidences that indeed the proposed regression model can serve as a strong discriminator to identify situations of loss of engagement (TD or BD in the UE-HRI coding scheme).
\end{enumerate}
Given these encouraging quantitative results, some qualitative assessment of exemplary frames with the corresponding computed engagement score are presented in figures~\ref{fig:engagement_}, \ref{fig:difficult_predictions} and \ref{fig:true_negative}. All figures show examples of the UE-HRI dataset, which was been completely absent from the training dataset (Section \ref{sec:TOGURO}).
Figure~\ref{fig:engagement_} presents two short sequences (roughly 2 seconds apart between frames), showcasing short-term diversion of attention of subjects resulting in a temporarily lower engagement score, but not leading to a very low engagement.
Figure~\ref{fig:difficult_predictions} exemplifies that our model can cope well with perception challenges which would forgo a correct assessment just using gaze or facial feature analysis. 
While one could in this context argue that our model has simply learned to detect people, figure~\ref{fig:true_negative} is providing three examples from different videos of the UE-HRI dataset with people present in the vicinity of the robot, but not engaging with it. 
The engagement score in these examples are significantly lower across all frames. 
We hypothesize that the learned model does not solely discriminate only person and/or face presence, but that the temporal aspects of the humans' behavior observable in the video are captured by the LSTM layer in our architecture well enough to successfully deal with these situations.

These qualitative reflections are evidently supported by the quantitative analysis on both datasets, providing us with confidence that the trained model is broadly applicable and can serve as a very useful tool to the HRI community with its modest computational requirements and high response speed in assessing videos from a robot's point of view. 

\begin{acks}
We thank the annotators, the Lincolnshire County Council and the museum's staff for supporting this research.
\end{acks}

\bibliographystyle{ACM-Reference-Format}
\bibliography{HRI20-engagement}

\end{document}